\documentclass[sigconf]{acmart}

\usepackage{hhline}

\AtBeginDocument{%
  }

\copyrightyear{2022}
\acmYear{2022}
\setcopyright{othergov}\acmConference[MMSports '22]{Proceedings of the 5th International ACM Workshop on Multimedia Content Analysis in Sports}{October 10, 2022}{Lisboa, Portugal}
\acmBooktitle{Proceedings of the 5th International ACM Workshop on Multimedia Content Analysis in Sports (MMSports '22), October 10, 2022, Lisboa, Portugal}
\acmPrice{15.00}
\acmDOI{10.1145/3552437.3555701}
\acmISBN{978-1-4503-9488-8/22/10}

\begin{document}

\title{KaliCalib: A Framework for Basketball Court Registration}

\author{Adrien Maglo}
\email{adrien.maglo@cea.fr}
\affiliation{\country{ }}

\author{Astrid Orcesi}
\email{astrid.orcesi@cea.fr}
\affiliation{%
  \institution{Université Paris-Saclay, CEA, List}
  \city{Palaiseau}
  \country{France}
  \postcode{F-91120}
}

\author{Quoc-Cuong Pham}
\email{quoc-cuong.pham@cea.fr}
\affiliation{\country{ }}

\renewcommand{\shortauthors}{Adrien Maglo, Astrid Orcesi, \& Quoc-Cuong Pham}

\begin{abstract}
  Tracking the players and the ball in team sports is key to analyse the performance or to enhance the game watching experience with augmented reality. When the only sources for this data are broadcast videos, sports-field registration systems are required to estimate the homography and re-project the ball or the players from the image space to the field space.
  This paper describes a new basketball court registration framework in the context of the MMSports 2022 camera calibration challenge. The method is based on the estimation by an encoder-decoder network of the positions of keypoints sampled with perspective-aware constraints. The regression of the basket positions and heavy data augmentation techniques make the model robust to different arenas.
  Ablation studies show the positive effects of our contributions on the challenge test set. Our method divides the mean squared error by 4.7 compared to the challenge baseline.
\end{abstract}



\begin{teaserfigure}
  \includegraphics[width=\textwidth]{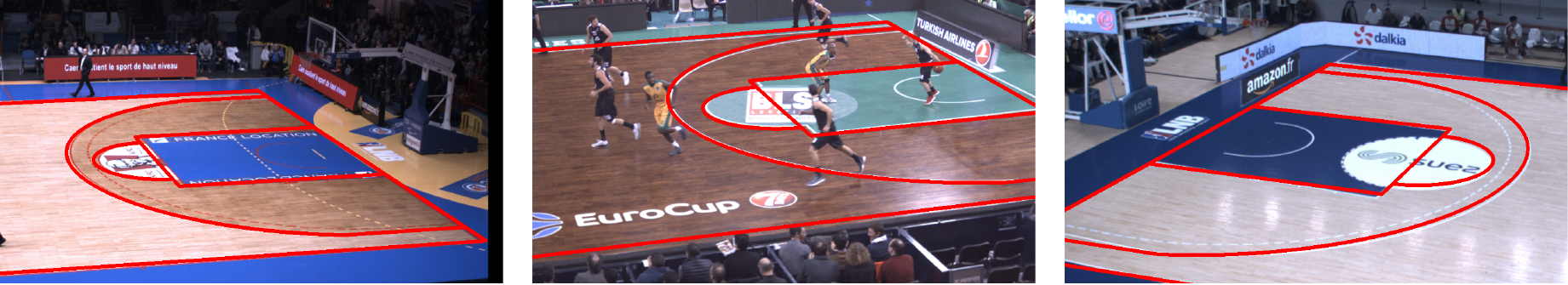}
  \caption{Baskeball court registration examples from the MMSPorts 2022 camera calibration challenge genearted by KaliCalib.}
  \label{fig:teaser}
\end{teaserfigure}

\maketitle

\section{Introduction}

The usage of data in team sport coaching is a growing trend. Player and ball positions are keys to the coaches for performance analysing, player scouting, train load monitoring or opponent analysis. Moreover, this data has also usage for entertainment. Indeed, virtual or augmented reality systems can enhance the viewing experience of a game by enabling replays with new point of views.

GPS tracking systems have been used for many years in outdoor sports. They are handy to use, even when a team is away his stadium. Yet, the coach can only gather data from his team and not the opponents. For indoors sports, tracking system exists but they require the arena to be specifically equipped.
The only readily available sources to determine the player and ball position are often broadcast videos. Thus, automatic player and ball tracking in team sports has been a growing research interest in the past years.

To compute the position of elements in the court space from a video, the location of the court should be known in each frame when the camera moves. Sports-field registration aims at estimating the homography between the 3D position of a sport-field and the 2D position in the image space.

Sports-field registration can be a difficult task. Sport fields are often composed of large homogeneous areas. The only available common distinctive features between them are the rule lines and their intersections. During the game, some of these features can be occluded by the players. When the camera moves quickly to follow the action, the field lines can also be blurry, which hinders their detection. In some sports, like basketball, the color design of the courts can vary a lot between arenas. Some logos and advertisements can be displayed on the court ground, thus changing their appearance.

For the MMSports 2022 workshop, a camera calibration challenge for basketball is organized \cite{zandycke2022deepsportradarv}. The aim is to retrieve the camera calibration parameters from single images of basketball games.
As, in the challenge, the calibration error is estimated only on the ground plane, we decided to estimate the calibration parameters with only points from this plane. This transforms the full calibration problem to a homography estimation.

We therefore propose KaliCalib, a new sports-field registration method based on the estimation by an encoder-decoder network of the positions of keypoints sampled with perspective-aware constraints. We also added the regression of the basket positions so as to help court localization. As the challenge training dataset was relatively small, we resorted to heavy data augmentation methods in order to generalize as much as possible to the test and challenge datasets.

In the first section of this paper, we review previous sports-field registration methods that process single images and available public sports-field registration datasets. In the second section, we describe the method and our contributions for the participation to the challenge. Finally, in the last section we perform ablative studies and describe the results we obtained on the test and challenge sets.

\section{Related work}

\subsection{Sports-field registration methods}

\begin{figure*}
  \centering
  \includegraphics[width=\linewidth]{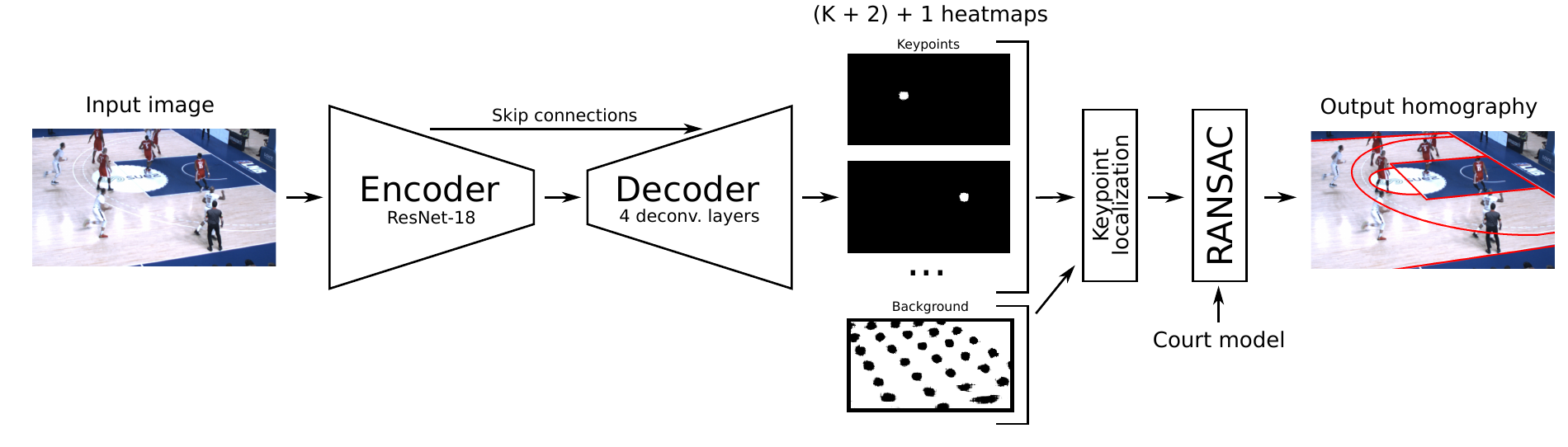}
  \caption{Overview of KaliCalib, our sports-field registration framework during inference. An encoder-decoder network generates keypoint and background heatmaps to retrieve the keypoint positions in the image space. A RANSAC algorithm then estimates the homography between the 3D positions of the court and the 2D positions in the frame. }
  \label{fig:kalicalib}
\end{figure*}

Some previous work focused on the sports-field registration for video with initial annotations \cite{hayet2004robust, okuma2004automatic, gupta2011using} or without \cite{wen2015court}.
In this section, we focus on fully automatic sports-field registration for single images as it is the target task of the MMSports 2022 camera calibration challenge \cite{zandycke2022deepsportradarv}.

The approach described in \cite{farin2003robust} uses a Hough transform to extract tennis court lines and then computes the homography with the line intersection points. Farin et al. \cite{farin2005fast} resorted to a RANSAC-like algorithm to detect the court lines.
The method of Yao et al. \cite{yao2017fast} detects intersections between soccer field lines and computes a feature vector to associate them to template points. Homography candidates are generated and then evaluated using a back projection to the frame space.
After having detected the soccer field lines and circles, the system of Cuevas et al. \cite{cuevas2020automatic} classifies them with a probabilistic decision tree. The homography is estimated with the intersection points between the classified elements.

Most of the recent approaches relies on a deep convolutional neural network to estimate the homography.
Homayounfar et al. \cite{homayounfar2017sports} uses a segmentation network to determine the play field surfaces, lines and circles. The matching between the images features and the template is formulated as a Markov random field solved by a branch and bound algorithm.
Sharna et al. \cite{sharma2018automated} matches the soccer field lines detected with pix2pix \cite{isola2017image} and a pre-built dictionary of synthetic edge maps. The homographies corresponding to the matched maps are then later refined with a Markov random field optimization.
In the approach of Sha et al. \cite{sha2020end}, the template matching is performed on segmented maps obtained with a U-Net network. Input image and template segmentations are transformed to features by a siamese network and compared with a \(L^2\) distance. The template homography is then refined by a spatial transformer network.
Ciopa et al. \cite{cioppa2021camera} train the same model with a teacher-student approach, the teacher being a commercial tool.
Chen and Little \cite{chen2019sports} use two GAN models : the first one to segment the soccer play field and the second one to extract the field lines. Then a siamese network matches the detected line features with an entry from a feature-pose database. The final homography is generated by refining the mapping of the detected and retrieved edges images thanks to a Lucas-Kanade algorithm \cite{baker2004lucas}.

Jiang et al. \cite{jiang2020optimizing} homography estimation framework is based on two models. The first network estimates the homography by regressing the real-world positions of four control points in the frame space. The second network estimates the re-projection error between the estimated field template and the real image. This allows to iteratively refine the initial homography estimation by minimizing this error. Fani et al. \cite{fani2021localization} also regresses four control point for hockey ice-rink registration.

The framework of Citaro et al. \cite{citraro2020real} computes the homography by estimating the position of court specific keypoints with a U-Net network.
Instead of using keypoint located on court corners, Nie et al. \cite{nie2021robust} sampled them on a regular grid. They also use a RANSAC algorithm to estimate the homography. The model is trained with a pixel classification approach. Chu et al. \cite{chu2022sports} recently improved the keypoint detection by using dynamic filters.

KaliCalib is built upon the approach of Nie et al. \cite{nie2021robust} without the homography refinement. We however resort to heavy data augmentation techniques, perspective aware keypoint sampling and basket localization to better fit the data of the MMSports 2022 camera calibration challenge.

\subsection{Public datasets}

Sports-field registration methods based on convolutional neural network need a significant amount of annotated images to be trained. Many authors resorted to private datasets to train and test their method.
Chen and Little \cite{chen2019sports} however publicly released their soccer WorldCup dataset. Composed of 395 images, it became the reference to compare recent sports-field registration approaches. Nevertheless, this small number of images can introduce biases in the training and evaluation of the appraoches. Chu et al. \cite{chu2022sports} therefore recently publicly released their TS-WorldCup datasets composed of 3812 soccer images.
For its second edition, the SoccerNet challenge \cite{deliege2021soccernet} also publicly released a soccer-field registration dataset with 20028 annotated frames.

There are therefore currently a significant offer of public datasets to train and evaluate soccer-field registration systems.
The situation is however not so satisfying for other sports. For the MMSports 2022 camera calibration challenge, the organizers released the first public dataset for basketball composed of 728 images.

\section{Proposed method}

\subsection{Overview}

KaliCalib estimates the position in the frame space of keypoints sampled on the field template thanks to an encoder-decoder convolutional neural network \cite{nie2021robust}. At inference time, this model generates heatmaps for each keypoint and the background. Keypoint positions are retrieved at the center of mass of the heatmaps. Since some of the positions may be inaccurate, a RANSAC algorithm \cite{fischler1981random} with DLT \cite{hartley2003multiple} estimates a robust homography between the court space and the image space by excluding outliers. This process is summarized on figure \ref{fig:kalicalib}.

\subsection{Network design}

KaliCalib's model generates heatmaps for each of the keypoints at a resolution of \(1/4\) of the input resolution. K different keypoints are sampled from the court template but the network also outputs a heatmap for the background that supervises the areas that do not contain any keypoint.

Our model has an encoder-decoder architecture similar to U-Net \cite{ronneberger2015u}. The encoder is based on a ResNet-18 network \cite{he2016deep}.
Dilated convolutions \cite{chen2017rethinking} with a factor two and non-local blocks \cite{wang2018non} are used in the last two ResNet blocks. Nie et al. \cite{nie2021robust} showed that they help to increase the receptive fields of local features. This improves keypoint localization in the case, for example, of occlusions or blur.
The decoder is composed of four deconvolution layers followed by a Relu activation. We also add skip connections between the encoder and the decoder to recover fine-grained details in the heatmaps.
The total number of parameters of our model is 14.5 millions.

\subsection{Training}

KaliCalib's network generates a softmax score for each pixel of the heatmaps that quantifies how much it corresponds to one keypoint or the background. It is trained with a cross-entropy loss:
\[
L = - \sum_{x} \sum_{k} \alpha_k \times p_{x,k} \times log(p_{x,k})
\]
where \(x\) is a pixel position at the heapmap resolution, \(k\) is a class (one of the keypoint or the background), \(\alpha_k\) is a constant weight for the class \(k\) and \(p_{x,k}\) is the softmax score of the heatmap pixel at position \(x\) for the class \(k\).

Instead of using single activated pixel for each keypoint on the ground truth heatmaps, we use a disk with a radius of 10 pixels to increase the number of pixels to positively regress.
Since the background class is still more represented than the keypoint classes, we set \(\alpha_k = 1000\) for the keypoint classes while \(\alpha_k = 1\) for the background class.

\subsection{Localizing the baskets}

The challenge input data provides the full camera calibrations parameters. As the dimensions of a FIBA basketball court are normalized, it is possible to retrieve the positions in the frame of the baskets. 
The baskets have very distinctive features that ease their localization.
Even if our approach estimates a homography between the court and the frame space, we train our model to generate two additional heatmaps for the baskets. These points cannot be used to estimate the homography but this supplementary task helps the network to localize correctly the field keypoints.

\subsection{Perspective aware keypoint sampling}

Nie et al. \cite{nie2021robust} regularly sample the keypoints in the court template.
With the effect of the perspective, the field keypoints far from the camera are more concentrated than the keypoints near the camera. Having a bigger keypoint concentration in the area far from the camera means that the RANSAC will more likely estimate the homography with keypoints from this area despite less pixels were available to the network to estimate the keypoint heatmaps.

\begin{figure}
  \centering
  \includegraphics[width=\linewidth]{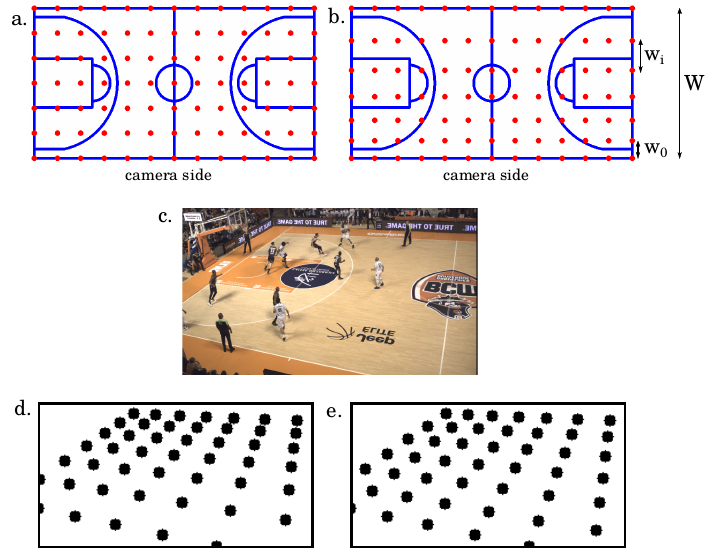}
  \caption{a. Uniform grid sampling of the keypoints. b. Perspective aware grid sampling of the keypoints. The points nearest to the camera are more spread out than the point farthest to the camera in order to compensate for the too large distance variations in the image generated by the perspective effect.
  c. Input image. d. Keypoints ground truth obtained with a uniform samping. e. Keypoints ground truth obtained with the perspective aware sampling.}
  \label{fig:samping}
\end{figure}

We therefore proposed a method to sample more uniformly the keypoints in the frame space, in the court width between the camera side and the opposite side. The differences between the two sampling method is shown on figure \ref{fig:samping}.
We model \(w_i\) the \(i^{th}\) distance between the keypoints in the court width axis and starting from the camera side by an arithmetic progression
\[
w_i = w_0 + i \times r
\]
with \(w_0\) the distance in the court width axis between the first two keypoints nearest to the camera and \(r\), the common difference.

Considering the sum of an arithmetic progression, we therefore have:
\[
W = (N - 1)\frac{2 \times w_0 + (N - 2) \times r}{2} \quad \textrm{for} \quad N \geq 2 
\]

with \(W\) the real-world width of the court and \(N\) the total number of keypoints on the court width axis.
We calculate \(r\) by setting a value to \(w_0\) with the following formula:
\[
r = \frac{2}{n - 2}({\frac{W}{n-1}} - w_0)
\]

We experimentally found that 1.75m was a good value of \(w_0\) for the dataset.

\subsection{Data augmentation}

To increase variety in training images, we use data augmentation techniques.
As in the training and test datasets, the images have been taken from the center of the court, it is possible to horizontally flip the images and the ground truth registration so the side of one team becomes the side of the other. Random image flips virtually increase the number of setups for each side.

Contrary to other sports such as soccer, the color appearance of basketball courts can be very different between arenas. In order to make our network more robust to the court color design, we resorted to heavy brightness, contrast and color jittering.

\section{Experiments}

\subsection{Dataset}

The DeepSportRadar Basketball Instants Dataset used in the challenge is composed of 480 images for the training, 164 for the validation, 84 for the testing and 84 for the challenge.
All the dataset images, except the challenge ones, come by pair, one for each side of the court. Random crops are extracted from both views to dynamically generate the train, validation and test datasets.

\begin{figure}
  \centering
  \includegraphics[width=\linewidth]{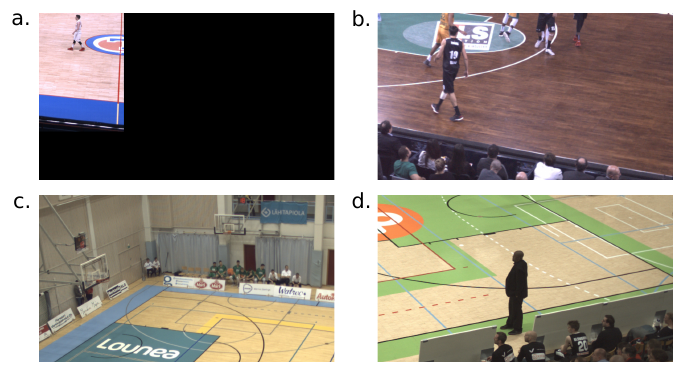}
  \caption{Hard examples from the MMSport 2022 camera calibration challenge test set (a. and b) and the challenge set (c and d). The visible area of the court is sometimes small in the test set. Multi-sports courts or arena with different configuration in the challenge set have many lines that can hinders the basketball court localization.}
  \label{fig:hard_dataset}
\end{figure}

It is quite challenging to learn robust basketball field registration with this dataset. The training set does not contain a wide variety of arenas. It is therefore hard to generalize to various court color designs. Besides, it is difficult to estimate the homography in some test and challenge images because only a small part of the court is visible or they contains lines from other courts as depicted on figure \ref{fig:hard_dataset}.

\subsection{Implementation details}

Our network was implemented with the Pytorch framework. It takes as input frames at the \( 960 \times 540\) resolution. We train it during 500 epochs with the AdamW optimizer, a batch size of 2 images and an initial learning rate of \( 10^{-4} \). After \( 2/3 \) of the epochs, the learning rate is divided by 10.
The original ResNet-18 encoder layers that have not been modified \cite{he2016deep} have been pre-trained on ImageNet \cite{russakovsky2015imagenet}. The non-local blocks of the encoder are initialized according to the Section 4.1 of the original paper \cite{wang2018non}. The decoder layers are initialized with default Pytorch uniform distributions.
We normalize the input images with the ImageNet mean and standard deviation values. The size of a FIBA basket court is 28m by 15m. The number of court keypoints \(K\) is set to 91. The RANSAC re-projection threshold is set to 35 pixels.

Regarding data augmentations, we use a random image flip probability of 0.5. For the color jittering, we use the torchvision \(ColorJitter\) transform with a brightness parameter of 0.7, a contrast parameter of 0.5, a saturation parameter of 0.5 and a hue parameter of 0.5.

In the case KaliCalib outputs a degenerated estimation or no homography, the plausible average homography of the training set is returned.
We detect degenerated homographies by re-projecting the frame points \([240,270]\) and \([720 ,270]\) in the court space and checking that their distance is below a threshold of 1800 cm.
The camera calibration parameters are computed with the estimated homography thanks to the OpenCV function \(calibrateCamera\) \cite{zhang2000flexible, bouguet2004camera}.

\subsection{Metric}

To evaluate the performance of our framework, we use the mean square error metric (MSE) as proposed by the challenge organizer. It is computed by re-projecting on the court space six frame points (left, center and right at the middle on bottom part of the frame).
The frame error is defined as the square root of the mean squared distance been the six points projected with the ground truth calibration and the estimated calibration.

\subsection{Results}

\begin{table}
    \begin{center}
    \begin{tabular}{|c|c|c|c|c|}
      \hline
      Color & Random & P.A. & Basket  & MSE  \\
      jittering & flip & sampling & local. & (cm) \\
      \hline
      x & x & x & x & 126.61 \\
      \hline
       & x & x & x & 154.93 \\
      \hline
      x & & x & x & 181.09 \\
      \hline
      x & x &  & x & 181.25 \\
      \hline
      x & x & x & & 130.68 \\
      \hhline{=====}
      \multicolumn{4}{|r|}{Nie et al. \cite{nie2021robust} - keypoints only} & 259.90 \\
      \hline
      \multicolumn{4}{|r|}{Challenge baseline} & 592.48 \\
      \hline
    \end{tabular}
    \end{center}
    \caption{Results on the test dataset of the MMSports 2022 camera calibration challenge. The result of Nie et al.'s method \cite{nie2021robust} was obtained with our implementation that does not contain the homography refinement.}
    \label{table:result}
\end{table}

A baseline approach was proposed by the challenge organizers. It is based on two models: the first one segments in the frame the 20 different court lines and the second one finds the line intersections and matches them with court points.
Our lowest MSE values for all the frames were respectively on the test and challenge sets 126.61 cm and 140.14 cm. For the test set, 76.5\% of the MSE measured on each image are inferior to one meter. The MSE for all the frames is divided by 4.7 compared to the baseline and by 2.1 compared to our implementation of Nie et al's method without the homography refinement. KaliCalib could not generate a homography for 2 images of the test set. An image is processed in about 9 ms on a system with an Intel Xeon Silver 4110 CPU and a NVIDIA GeForce GTX 1080 Ti GPU.

We performed an ablation study to measure the improvement of our contributions. The results, presented on table \ref{table:result} shows the positive impact of our contributions.
The color jittering reduces the error by about 18\%, random image flips by about 30\%,
the perspective aware sampling of the keypoints by about 30\% too and the basket localization by about 3\%.

\section{Conclusion}

We presented in this paper KaliCalib, a new sports-field registration framework we developed to participate to the MMSports 2022 camera calibration challenge. On the challenge test set, KaliCalib divides the mean squared re-projection error by 4.7 compared to the baseline.
As future work, we would like to test this approach on other datasets and study new network architectures to better capture the field features.

\begin{acks}
This work benefited from a government grant managed by the French National Research Agency under the future investment program (ANR-19-STHP-0006) and the FactoryIA supercomputer financially supported by the Ile-de-France Regional Council.
\end{acks}

\bibliographystyle{ACM-Reference-Format}
\bibliography{main}

\end{document}